\documentclass[a4paper,fleqn]{cas-dc}

\usepackage[numbers, compress]{natbib}

\usepackage{algorithm}
\usepackage{algpseudocode}
\usepackage{caption}

\begin{document}
\let\WriteBookmarks\relax
\def\floatpagepagefraction{1}
\def\textpagefraction{.001}

\shorttitle{USL-Net: Uncertainty Self-Learning Network for Unsupervised Skin Lesion Segmentation}

\shortauthors{Xiaofan Li et al.}

\title [mode = title]{USL-Net: Uncertainty Self-Learning Network for Unsupervised Skin Lesion Segmentation}

%

\author[1]{Xiaofan Li}[orcid=0000-0002-5011-6314]


\ead{xfl@my.swjtu.edu.cn}

\credit{Conceptualization, Methodology, Writing-original draft, Writing-reviewer and editing}

\author[1,2,3,4]{Bo Peng}
\cormark[1]

\ead{bpeng@swjtu.edu.cn}

\credit{Writing-original draft, Writing-reviewer and editing, Funding acquisition, Project administration.}

\affiliation[1]{organization={School of Computing and Artificial Intelligence, Southwest Jiaotong University},
            city={Chengdu},
            postcode={611756},
            state={P.R},
            country={China}}

\affiliation[2]{organization={Engineering Research Center of Sustainable Urban Intelligent Transportation, Ministry of Education},
            city={Chengdu},
            postcode={611756},
            state={P.R.},
            country={China}}
\affiliation[3]{organization={National Engineering Laboratory of Integrated Transportation Big Data Application Technology, Southwest Jiaotong University},
            city={Chengdu},
            postcode={611756},
            state={P.R.},
            country={China}}
\affiliation[4]{organization={Manufacturing Industry Chains Collaboration and Information Support Technology Key Laboratory of Sichuan Province, Southwest Jiaotong University},
            city={Chengdu},
            postcode={611756},
            state={P.R.},
            country={China}}

\affiliation[5] {organization={Data Recovery Key Laboratory of Sichuan Province, Neijiang Normal University}, 
    city = {Neijiang,Sichuan},
    postcode={641100},
    state={P.R.},
    country={China}}

\cortext[1]{Corresponding author. School of Computing and Artificial Intelligence, Southwest Jiaotong University, Chengdu, 610031, China}


\author[1,2,3,4]{Jie Hu}
\ead{jiehu@swjtu.edu.cn}
\credit{Methodology, Project administration.}

\author[5]{Changyou Ma} 
\ead{scdr@njtc.edu.cn}
\credit{Funding acquisition, Project administration.}

\author[1]{Daipeng Yang}
\ead{daipengyang@my.swjtu.edu.cn}
\credit{Methodology, Writing-original draft.}

\author[1]{Zhuyang Xie}
\ead{zyxie@my.swjtu.edu.cn}
\credit{Conceptualization, Methodology.}


\begin{abstract}
Unsupervised skin lesion segmentation offers several benefits, such as conserving expert human resources, reducing discrepancies caused by subjective human labeling, and adapting to novel environments. However, segmenting dermoscopic images without manual labeling guidance is a challenging task due to image artifacts such as hair noise, blister noise, and subtle edge differences. In this paper, we introduce an innovative Uncertainty Self-Learning Network (USL-Net) to eliminate the need for manual labeling guidance for the segmentation. Initially, features are extracted using contrastive learning, followed by the generation of Class Activation Maps (CAMs) as saliency maps. High-saliency regions in the map serve as pseudo-labels for lesion regions while low-saliency regions represent the background. Besides, intermediate regions can be hard to classify, often due to their proximity to lesion edges or interference from hair or blisters. Rather than risking potential pseudo-labeling errors or learning confusion by forcefully classifying these regions, they are taken as uncertainty regions by exempted from pseudo-labeling and allowing the network to self-learning.
Further, we employ connectivity detection and centrality detection to refine foreground pseudo-labels and reduce noise-induced errors. The performance is further enhanced by the iterated refinement process. The experimental validation on ISIC-2017, ISIC-2018, and PH2 datasets demonstrates that its performance is comparable to supervised methods, and exceeds that of other existing unsupervised methods. On the typical ISIC-2017 dataset, our method outperforms state-of-the-art unsupervised methods by 1.7\% in accuracy, 6.6\% in Dice coefficient, 4.0\% in Jaccard index, and 10.6\% in sensitivity.
\end{abstract}



\maketitle

\section{Introduction}\label{sec1}

Skin cancer is a global health concern, with millions of new cases reported annually. Early detection and accurate diagnosis of skin lesions are critical for successful treatment and improving patient outcomes. Historically, the interpretation of skin lesions heavily depended on the subjective judgment of medical professionals, often resulting in substantial variations \cite{b1}. The advent of artificial intelligence (AI) has facilitated objective, intelligent analysis of skin lesion images using machine algorithms. In the context of AI-assisted dermatology, skin lesion image segmentation is a fundamental technique, which ensures accurate delineation of lesion boundaries for subsequent analysis and diagnosis. It generally refers to dividing an image into cancerous regions (foreground) and other regions (background).

The progress of deep learning techniques, such as convolutional neural networks (CNNs), has significantly enhanced image segmentation qualities. Networks such as FCN \cite{b4}, U-Net \cite{b5}, and DenseNet \cite{b6} are successfully applied to skin lesion segmentation, leading to a substantial improvement in the performance of both supervised and weakly supervised methods. Supervised methods \cite{b7} required manual labeling of a large number of pixel-level labels for learning, while weakly supervised methods \cite{b9} involved labeling at the image level or with less data. Recently, some weakly supervised skin lesion segmentation methods \cite{b52}\cite{b53} achieved superior results by attempting to produce pseudo-labels from thresholding the class activation map (CAM) \cite{b43}. They have the limitation of biasing towards notable features rather than the lesion regions only. Therefore disturbing factors such as hair and air bubbles will obstruct the segmentation process. Moreover, directly thresholding the CAM can result in issues such as over-segmentation or under-segmentation. Both of the supervised and the weakly supervised approaches demand a considerable amount of professional input to generate the necessary annotations. Although weakly supervised methods can reduce the workload to some extent, they still rely on manually labeled data. Furthermore, these methods are not directly applicable to new scenarios without re-labeling the new data.

The emergence of unsupervised skin lesion segmentation offers a promising solution to the above challenges. Traditional unsupervised methods involve the analysis of texture and color disparities between the lesion region and the background. The segmentation was performed by using a binary threshold \cite{b13}, or using region-growing algorithms \cite{b14} for skin lesion segmentation. These approaches differ significantly from deep learning methods in terms of performance. It is primarily due to the presence of noise in dermoscopic images, as well as the inadequacy of relying on basic attributes such as texture and color.
For deep learning based methods, many pre-trained models \cite{b39,b40} were proposed based on contrastive learning. Without manual labels, pseudo-labels were produced from the networks. In contrast to biasing towards only on notable features, contrastive learning focuses on distinguishing instances and produces the CAM with high saliency for the entire instances. We observe that for skin lesion segmentation, the class labels can be obtained by generating a CAM using contrastive learning, since there is only one class of instances and only one instance in an image (lesion region). This can offer a potential solution to this challenge.

\begin{figure}[!htbp]
\begin{center}
\includegraphics[width=0.9\linewidth]{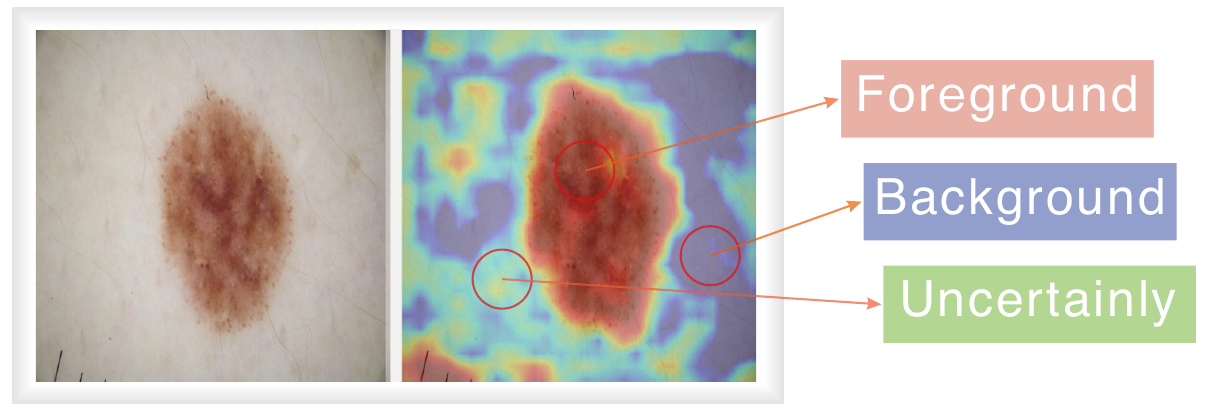}
\caption{A reference example of our idea in splitting classes based on CAM as a saliency map. Red color means higher saliency, green represents the region of medium saliency and blue represents the region of low saliency.}
\label{figure1}
\end{center}
\end{figure}

Based on this idea, we introduce a novel unsupervised skin lesion segmentation method called Uncertainty Self-Learning Network (USL-Net).
 The network employs contrastive learning to extract image features. It can reduce the dependency on manual labeling and accurately locate the lesion instance. Subsequently, CAMs are generated based on these features and thereby produce pseudo-labels. Figure.\ref{figure1} provides an example of labeling the CAM. On the left is the original image, and on the right is the CAM. Regions in low saliency, depicted in \textbf{blue}, can confidently be classified as the background. While regions with high saliency, shown in \textbf{red}, can be identified as the foreground after applying techniques such as connectivity detection \cite{b99} to filter unique regions. To determine the class of medium saliency regions (denoted in \textbf{green}), especially those near the edges of the lesion region, we consider them as uncertain locations and classify them by USL-Net. Additionally, the foreground pseudo-labels will be further refined based on the uniqueness of lesions.
The USL loss function is proposed to guide the learning process, which disregards uncertain locations and enables the network to autonomously learn from them. Moreover, an iterative-refinement process is introduced to further enhance the overall quality of the pseudo-labels.

Our main contributions can be summarized as follows:
\begin{itemize}
  \item An unsupervised self-learning network (USL-Net) is proposed to learn uncertainty regions, which effectively avoids some pseudo-labeling errors due to the threshold setting. The unsupervised learning strategy can largely reduce the dependency on manual labeling and improve lesion segmentation performance.
  \item An uncertainty module (UM) is designed to effectively filter uncertainty locations and ensure high-quality pseudo-labels. Additionally, we present an iterative-refinement process to enhance segmentation results for lesion region prediction.
  \item Comprehensive experiments are conducted on ISIC-2017 \cite{b20}, ISIC-2018 \cite{b21} and PH2 \cite{b22} datasets with various validation criteria and comparative methods, which demonstrate the effectiveness of the proposed method.
\end{itemize}

The rest of the paper is organized as follows: Section 2 briefly reviews the most related methods in the literature. Section 3 provides a detailed explanation of the proposed method. In Section 4, the experimental results are presented with ablation and a comparison to state-of-the-art methods. Section 5 discusses the improved effectiveness and rationalization of our approach, and Section 6 presents the conclusion.

\section{Related work}
\label{sec2}

\subsection{Skin lesion segmentation}

Traditional skin lesion segmentation methods are mainly based on unsupervised learning, which often relies on manually designed features, such as boundary \cite{b33}, contour \cite{b34} and colors \cite{b11} to identify the lesion areas. In boundary or contour-based approaches, earlier work from Yuan et al. \cite{b33} treated skin lesion segmentation as a search problem, utilizing evolution strategies to locate lesion region boundaries. R. Kasmi, et al. \cite{b34} proposed a geodesic active contouring technique for skin lesion segmentation. In the context of color space, Sagar et al. \cite{b11} presented a binary thresholding method which examined the color differences between the lesion region and the background. Louhichi et al. \cite{b13} extracted features from three color spaces (RGB, HSV and XYZ), and integrated the multiple density clustering and the region-growing algorithms for skin lesion segmentation. Ramya et al. \cite{b37} removed artifacts like ink and color cards using the discrete wavelet transform and differentiated lesion regions from the background through threshold-based Otsu and histogram methods. Confined to the feature representation capability, traditional machine learning methods can not fully describe the complex textures, colors or shapes information, therefore can not achieve good performance in lesion segmentation.

Many deep learning methods have been proposed for medical image processing \cite{b114}\cite{b115}. In skin lesion segmentation, CNN-based networks such as U-Net, FCN and VGG are widely studied. For instance, iFCN \cite{b113} extended the FCN by incorporating the residual structure from ResNet \cite{b54}. Lin et al. \cite{b24} demonstrated the superiority of U-Net through a comparison with C-means clustering methods. AS-Net \cite{b8} employed the VGG network as the backbone and strengthened it with an attention module as well as a synergistic module to integrate spatial and channel attention path features. Despite the high accuracy achieved, these methods are completely data-driven which often require a substantial amount of time for intensive pixel-level labeling.  Due to the large dependence on the data, they are difficult to be generalized in new scenarios.

There are methods that attempt to alleviate the expensive labeling associated with supervised segmentation. SWSDB model \cite{b52} combined two networks: the coarse-SN network used a small number of real segmentation labels to learn, and the dilated-CN network based on image-level category labels to generate CAMs. Then they are fed jointly into the enhanced-SN network to produce the final segmentation. VitSeg \cite{b53} roughly identified the lesion region through the CAM, then carried out saliency propagation to yield the segmentation result. Since the CAM only provides a coarse description of the lesion area, further processing is required to reduce the potential noise and improve the final result.

\subsection{Contrastive learning and CAM}

Contrastive learning is a self-supervised discriminative method, where manual labels are not required. Instead, labels are produced based on the intrinsic properties of the data itself \cite{b38}, making most contrastive learning methods fall under the unsupervised field. Methods like SimCLR \cite{b39}, MoCov1 \cite{b40}, and MoCov2 \cite{b41} used augmentations of an original image as the positive samples and those of other images as the negative samples. Transformer-based methods, such as MoCov3 \cite{b111}, DINO \cite{b112}, and DINOv2 \cite{b116} are consistent in designing positive and negative sample pairs. However, it is worth noting that current contrastive learning methods exhibit high sensitivity to noise and uncertainty, potentially leading to the model learning inaccurate representations. To enhance stability, it may be necessary to combine multiple techniques. Furthermore, these methods are typically employed for learning generic feature representations rather than being optimized for specific tasks. As a result, further design is often needed for subsequent stages tailored to specific problems.

The CAM is a heatmap generated based on the feature space during network learning, highlighting deep feature clues such as target or instance classes. The key distinction between unsupervised and weakly supervised CAM methods lies in the acquisition of feature maps. Common methods to obtain class activation maps include CAM \cite{b42}, Grad-CAM \cite{b43}, Grad-CAM++ \cite{b44}, Smooth Grad-CAM++ \cite{b19}.
Some recent improvements such as FreqCAM \cite{b99} set a threshold based on frequency information and employed connectivity search to filter the largest region as the instance region. CCAM \cite{b45}, on the other hand, focused on retraining to distinguish the foreground and the background based on contrastive learning. It generated class-agnostic activation maps that do not differentiate between foreground classes. These advancements provide valuable insights for improving the quality of saliency maps, which align well with the requirements for skin lesion segmentation.

\section{Proposed Method}
Existing weakly supervised methods in the field usually follow a three-step process: 1) Feature extraction is performed; 2) CAM is generated in the feature space, and pseudo-labels are directly obtained using binary thresholding on CAM; 3) Learning is conducted with these pseudo-labels. To ensure the reliability of pseudo-labels and classify all pixels correctly, these methods often require fine-tuning with some ground-truth annotations. In this study, we first explore to extract rich features from a fusion of three contrastive learning methods (section \ref{sec:contrastive_learning}).  Observing that the class activation maps usually show higher saliency for lesion regions and lower saliency for background regions. In the between, there are still some lesion regions that are shown in the medium saliency, especially the marginal regions. Moreover, there are some background regions (noisy regions such as hair, blisters, etc.) that present in the medium or even high saliency. These medium-saliency regions can hardly be removed by connectivity and centrality methods or simply thresholding them out. Along the line of learning, we treat them as uncertainty regions, allowing the network to learn their labels based on contextual cues (sections \ref{sec:CAM}, \ref{sec:UM}, \ref{sec:Optimize}). This strategy aims to alleviate the potential pseudo-label errors during network learning. Finally, the trained encoder-decoder network will be used for inferring the segmentation result (section \ref{sec:inference}).

  \begin{figure*}[htbp]
\begin{center}
\includegraphics[width=0.9\linewidth]{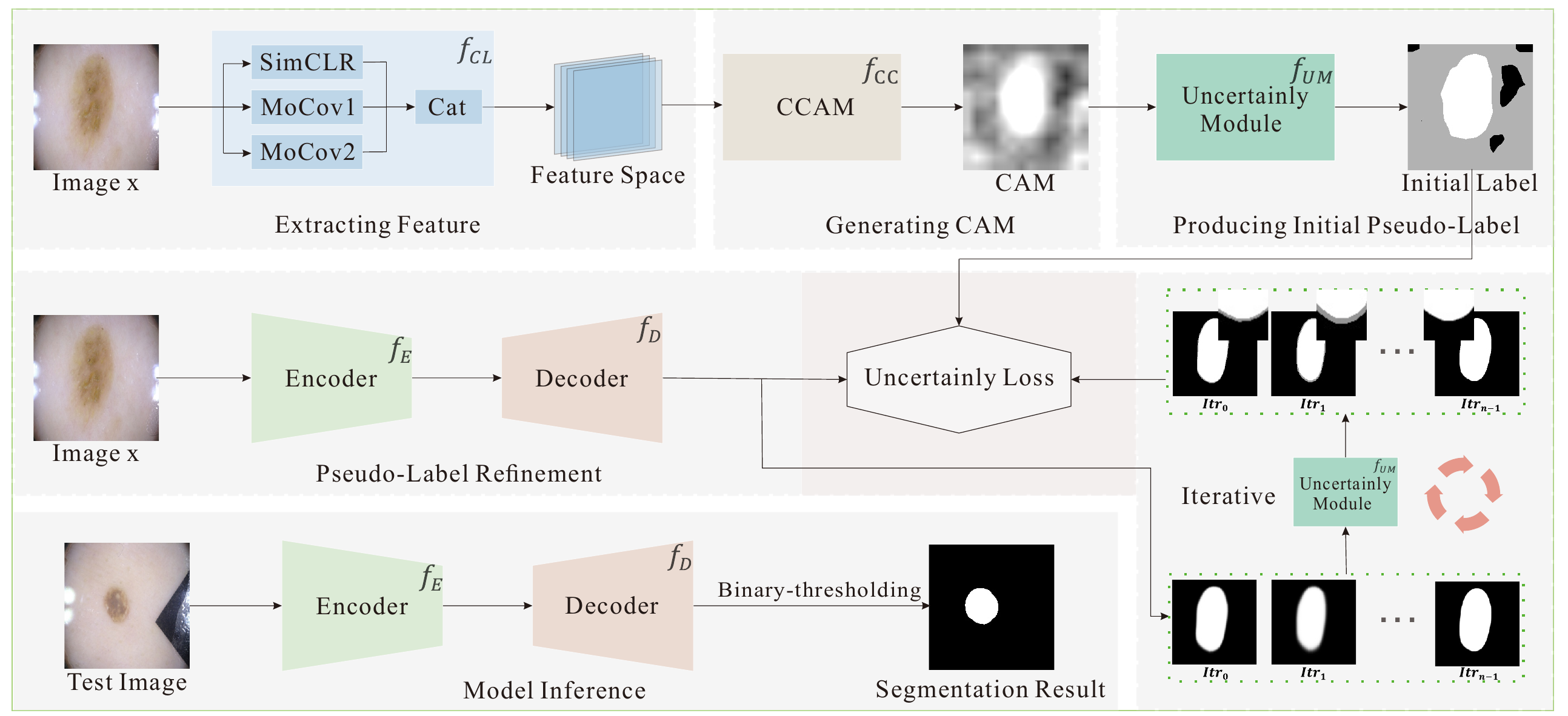}
\caption{USL-Net consists of a fusion of multiple contrastive learning module for feature extraction and a CCAM module for generating the CAM. The Uncertainty Module (UM) is responsible for converting CAMs into pseudo-labels. The encoder and decoder modules learn based on the pseudo-labeling with the proposed uncertainty loss, and the results can be used as new pseudo-labeling. During the inference phase, segmentation results are directly generated using binary thresholding after the data passes through the encoder and decoder modules.}
\label{figure2}
\end{center}
\end{figure*}

\subsection{Overview of the proposed USL-Net} \label{sec:USL-NET}
The structure of USL-Net is depicted in Figure \ref{figure2}. For a given image $x$, we use contrastive learning denoted as $f_{CL}$ to extract image features. Specifically, SimCLR, MoCov1 and MoCov2 stand for different contrastive learning methods (see section 3.2), and Cat stands for the concatenation operator. We apply the CCAM (see section 3.3), denoted as $f_{CC}$, to generate CAM, which we define as $C$. Consequently, the CAM can be expressed as $C = f_{CC}( f_{CL}(x) )$, where $x$ is in $h \times w \times c$ and $C$ is $h \times w$. In our study, the values for $h$, $w$, and $c$ are set as defaults to be $224$, $224$ and $3$, respectively.

To generate the initial pseudo-labels $l_0$, we utilize $C$ as the input to obtain $l_{0}$ by the uncertainty module $f_{UM}$, i.e., $l_{0}=f_{UM}(C)$. The size of $l_0$ and $C$ are the same, with each element $l_{0_{i, j}} \in [0, 0.5, 1]$, where $0\leq i<h$ and $0\leq j<w$. The values of $0$, $1$ and $0.5$ represent background, foreground and uncertainty locations, respectively.

Once $l_0$ is obtained, we generate a prediction result $p$ by feeding the image $x$ into the encoder and decoder networks $f_E$ and $f_D$, respectively. In other words, we have $p = f_D(f_E(x))$ with the same size as $l_0$. Furthermore, we use pseudo-label $l_0$ as a reference to calculate the foreground and background differences between $p$ and $l_0$, and employ the uncertainty loss function to optimize the learning process.

Compared with the original CAM result, the outcomes produced by USL-Net offer a better distinction between the foreground and the background. Furthermore, we introduce an iterative-refinement module, where the generated results are sent back into the Uncertainty Module (UM) for the next iteration of generating pseudo-labels. In the inference phase, skin lesion images are sent into the encoder and decoder modules, and the segmentation results are obtained through binary thresholding. Note that both of contrastive learning and CCAM are integral components of a pre-learning process, which are directly applied in the extraction of pseudo-labels.

\subsection{Feature extraction} \label{sec:contrastive_learning}
The feature extraction is performed by a contrastive learning process, which aims to discriminate the positive and the negative sample pairs. To collect rich features and stabilize the learning results, we construct a contrastive learning fusion module, which can accommodate various contrastive learning structures rooted in convolutional neural networks. Typically, we choose three popular contrastive learning methods, i.e, A Simple Framework for Contrastive Learning of Visual Representations (SimCLR), Momentum Contrast (MoCov1), and the Improved Baselines with Momentum Contrastive Learning (MoCov2). They are fused with ResNet50 \cite{b54} as the backbone network. In the experiments, we will assess and compare the performance of the CCAM-based method with other structures such as transformer based methods in generating class activation maps.

Specifically, the input image is fed into the ResNet50 network for feature extraction. Note that we make a modification to the last convolution stride of the $5^{th}$ layer, setting it to $1$. This modification is aimed at retaining spatial features and minimizing the loss of spatial information. Afterward, the extracted features consist of the ResNet50 layer $4$ and layer $5$ features.

\begin{figure}[htbp]
\begin{center}
\includegraphics[width=0.9\linewidth]{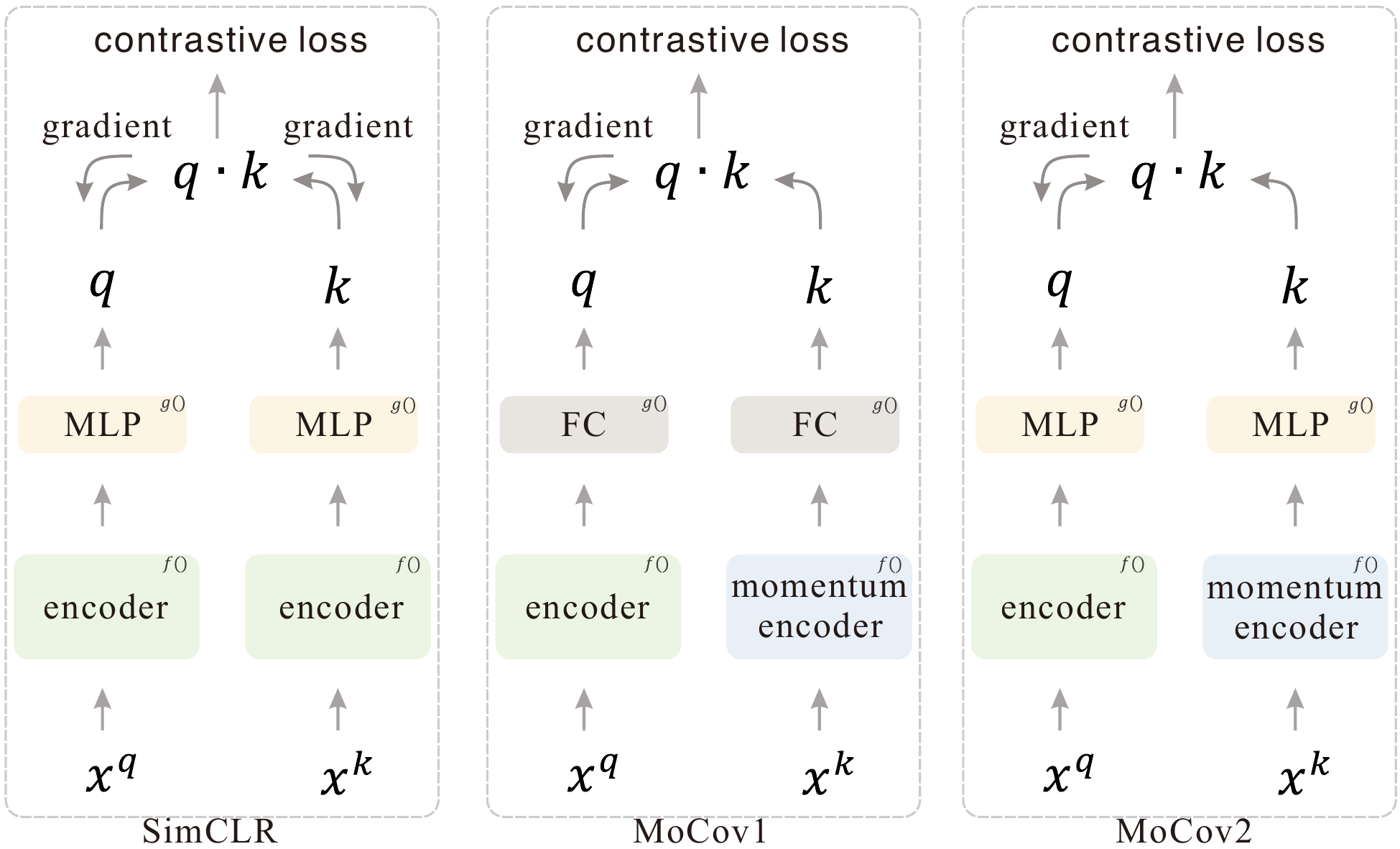}
\caption{The learning process of three contrastive learning methods. The main differences between the three methods are that MoCov1 and MoCov2 use a momentum decoder and the gradient is unidirectional. SimCLR and MoCov2 use MLP in the feature mapping layer, and MoCov1 uses FC.}
\label{figure4}
\end{center}
\end{figure}

Figure \ref{figure4} provides a visual illustration of the contrastive learning methodology. For a given image $x^q$, we define $x^k$ to include all augmented images within the same batch. $x^{k_+}$ is defined as the augmentation of $x^q$, whereas $x^{k_-}$ represents other augmentations in $x^k$. The contrastive learning process aims to minimize the feature distance between $x^{k_+}$ and $x^q$ and maximize the distance between $x^{q}$ and other augmentations $x^{k_-}$.

In the SimCLR method, image augmentations serve as the positive samples for the image itself, while augmentations of other images in the same batch act as negative samples. Especially, both $x^q$ and $x^k$ in the SimCLR undergo the backward propagation.
In contrast, the MoCov1 method employs a memory bank to store the feature representations of all samples in the $x^k$ branch (stored within the momentum encoder module) and randomly selects some of these as negative samples for the learning process. This branch does not participate in the backward propagation.
Building on the foundation of MoCov1, MoCov2 utilizes Multilayer Perceptron (MLP) to output feature vectors and includes blur augmentation in the process. It is worth noting that SimCLR also makes use of an MLP, while MoCov1 utilizes a Fully Connected Layer (FC).

All the three contrastive learning methods uniformly employ InfoNCE as the contrastive loss function. For a given input $x^q$, the InfoNCE loss, denoted as $L_{x^q}$, is defined in Equation.\eqref{eq1}. $x^{k_i}$ represents the $i^{th}$ image's augmentation in $x^k$, $K$ denotes the number of negative samples, and $\tau$ is a hyperparameter, typically setting to $0.07$.

\begin{equation}
L_{x^q}= -\log\frac{\exp (x^q \cdot x^{k_+}/\tau)} {\sum^K_{i=0} \exp (x^q \cdot x^{k_i}/\tau)}
\label{eq1}
\end{equation}

\subsection{Generating the class activation map (CAM)} \label{sec:CAM}
\begin{figure}[htbp]
\begin{center}
\includegraphics[width=0.9\linewidth]{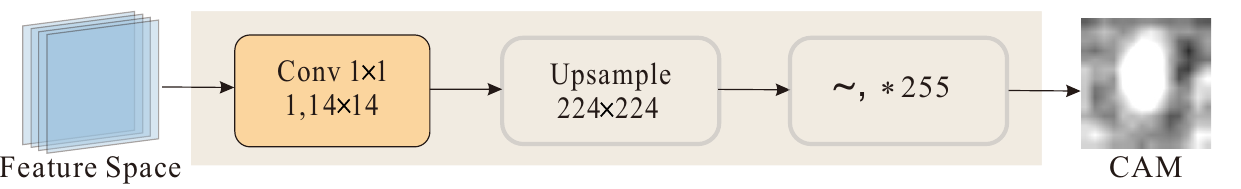}
\caption{The structure of CCAM for generating CAM.  It contains a convolutional layer, an upsampling layer, and an inverse operation.}
\label{figure5}
\end{center}
\end{figure}

In ResNet50,  features extracted from the $4^{th}$ and $5^{th}$ layers are concatenated and utilized as input for generating the class activation map (CAM) via the CCAM. The CCAM consists of a $1 \times 1$ convolution module, an upsampling module, and a product operation. Figure \ref{figure5} presents an illustration of the CCAM structure.

Feature maps $F$ are extracted via contrastive learning. $F$ is reshaped into $F_{fla}$, with the dimensions of $4096 \times 256$. We obtain a background class-agnostic activation map $M$ by convolving $F$ with an $1 \times 1$ operator. And $M'$ as the foreground class-agnostic activation map is calculated by the complement operation on $M$. We utilize bilinear interpolation to up-sample $M$ to obtain a $224 \times 224$ map. After scaling it by 255, we get the CAM map $C$.

Let $F_{T}$ be the transpose of $F_{fla}$. The map $M'$ is reshaped to $M'_{fla}$ in the dimensions of $4096 \times 1$, and similarly, $M$ is reshaped to $M_{fla}$ in the dimensions of $4096 \times 1$. We then multiply $F_{T}$ with both $M_{fla}$ and $M'_{fla}$, thus yielding feature representations for the background and foreground respectively, each with the dimensions of $256 \times 1$. The aim of this learning process is to create discrete representations for the foreground and the background.

Among various multi-class activation map generation methods available, CCAM is designed as a relearning process for distinguishing the foreground and the background. This characteristic aligns well with the requirement for skin lesion segmentation.

\begin{figure}[htbp]
\begin{center}
\includegraphics[height=4cm,width=8.8cm]{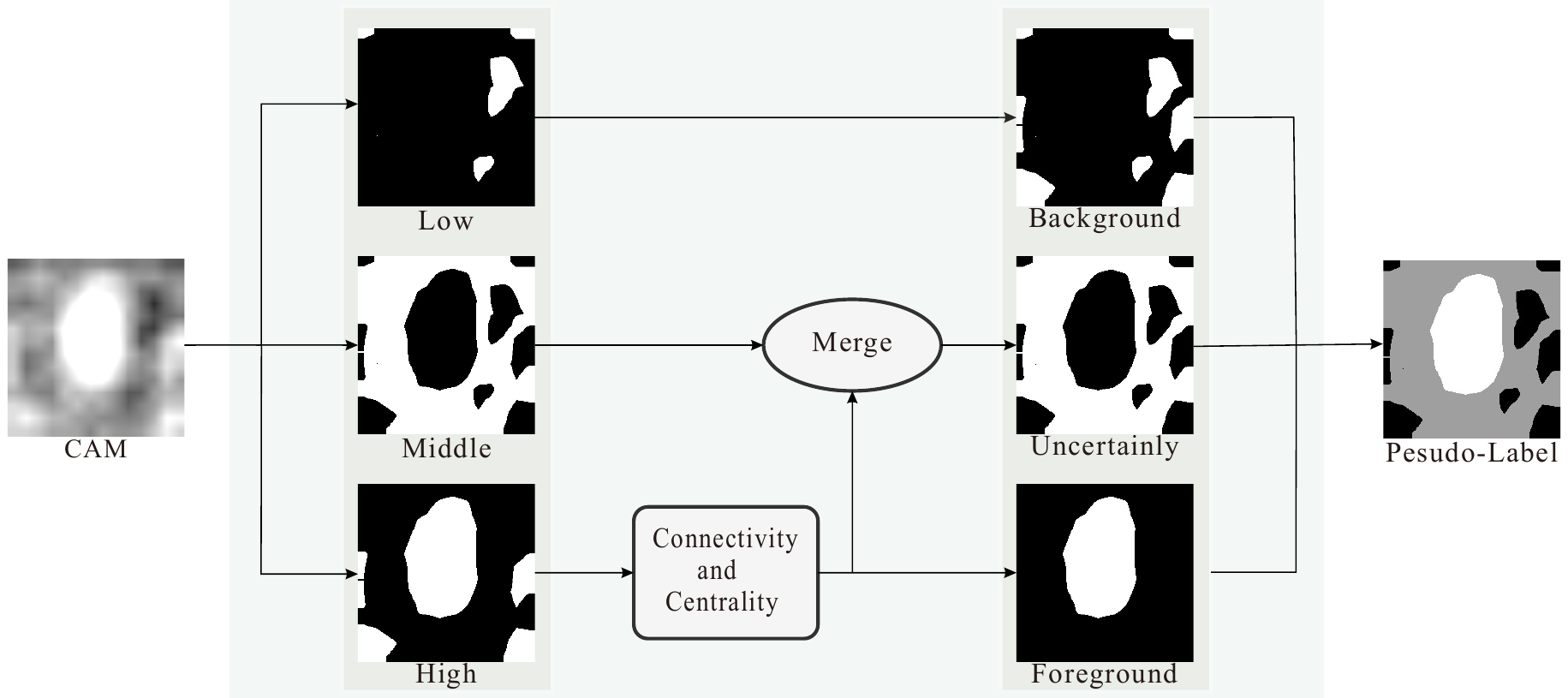}
\caption{Producing pseudo-label by uncertainty module.}
\label{figure6}
\end{center}
\end{figure}

\renewcommand{\algorithmicrequire}{\textbf{Input:}}
\renewcommand{\algorithmicensure}{\textbf{Output:}}

\begin{algorithm}[h]
    \caption{Connectivity and centrality  filtering}
    \label{alg1}
      \begin{algorithmic}[1]
           \Require High-saliency map $C_{h}$, the height H and the width W for $C_{h}$
           \Ensure Foreground map  $F_{map}$
           \State $ RP \gets$ Random\_Point($C_{h}$)
           \State $n \gets$ 0
           \While{True}
               \State $region_n \gets Connectivity\_Search(C_h, RP)$
               \If{all locations are searched}
                                  \State   Stop loop
               \EndIf
               \State $RP\gets$ Random\_Point($C_{h}$) and $RP \notin region_{0,...n}$
               \State $n++ $
           \EndWhile
           \State $weight \gets Centrality(region_0, h, w)$        \Comment eq(2)
           \State $k \gets 0$
           \State $\alpha \gets 500$
           \For{i  \textbf{in} [1,...,n]}
           {
               \If{$len(region_i) < \alpha$ }
                  \State $C_{h}[region_i] \gets 0 $
               \ElsIf{$Centrality(region_i, H, W) > weight$}
                   \State $weight \gets Centrality(region_i, H, W)$
                   \State $k \gets i$
               \EndIf
           }
           \State $F_{map} \gets C_{h}[region_k] $
           \EndFor
          \State \Return $F_{map}$

       \end{algorithmic}
\end{algorithm}

\subsection{Producing initial pseudo-labels with Uncertainty module (UM)} \label{sec:UM}

The structure of UM is shown in Figure \ref{figure6}. The image is first divided into three parts. Regions with low saliency are explicitly defined as the background. Regions with medium saliency are designated as uncertainty regions. While regions with high saliency are further classified into foreground and uncertainty regions using connectivity and centrality detection. We set threshold $0.65$ for the high saliency regions, and $0.35$  for the low saliency regions. Values in the between are classified as the medium saliency regions.

The connectivity and centrality algorithms are devised to ensure the reliability of foreground pseudo-labeling. By analyzing the skin lesion data, we observe two key properties: (1) The lesion regions are unique and densely interconnected; (2) In cases where multiple saliency regions exist, the regions close to the image boundaries are often associated with black shadow noise rather than true lesion regions. Consequently, for the high saliency regions, we partition them using the connectivity detection method and filter out regions that are excessively small. Following that, we identify the unique regions that are proximal to the center and distant from the edges using the centrality detection method.

We illustrate the method in Algorithm \ref{alg1}. The input is a map  $C_{h}$, which only contains high-saliency regions, and all other regions are assigned to $0$s. RP is a randomly selected point within $C_{h}$, and the function $Connectivity_Search()$ implements the Flood-Fill algorithm. A predetermined hyper-parameter $\alpha$ is utilized to filter out regions that are too small. In this study, we set $\alpha=500$. The function $len()$ is used to determine the number of points in a region. Afterward, $Centrality()$, the centrality filter function, is calculated using Equation \eqref{eq2}. For the $i^{th}$ connected region in $C_{h}$, $H$ and $W$ represent the height and width of $C_h$, respectively. The sum of the distances to the center of each region equates to the $weight$. The region with the minimum weight is labeled as the foreground, while the remaining regions are considered as uncertain locations. It is important to note that if there is no identifiable foreground after filtering or if the foreground is located too far from the center, the sample is excluded.

\begin{equation}
weight= \sum{  \sqrt {(region_{i}-\frac {H}{2})^2+(region_{i}-\frac {W}{2})^2}}
\label{eq2}
\end{equation}

\subsection{Pseudo-label Refinement} \label{sec:Optimize}
To facilitate pseudo-label supervised learning, we construct a backbone architecture using ResNet50 coupled with a Fully Convolutional Network (FCN). FCN is widely utilized in deep learning-based image segmentation. It integrates both convolutional and bilinear interpolation operations in the encoder (Figure \ref{figure2} Encoder) and decoder stages (Figure \ref{figure2} Decoder) to maintain the consistency between the sizes of the output and input.

\begin{figure*}[h]
\begin{center}
\includegraphics[width=0.8\linewidth]{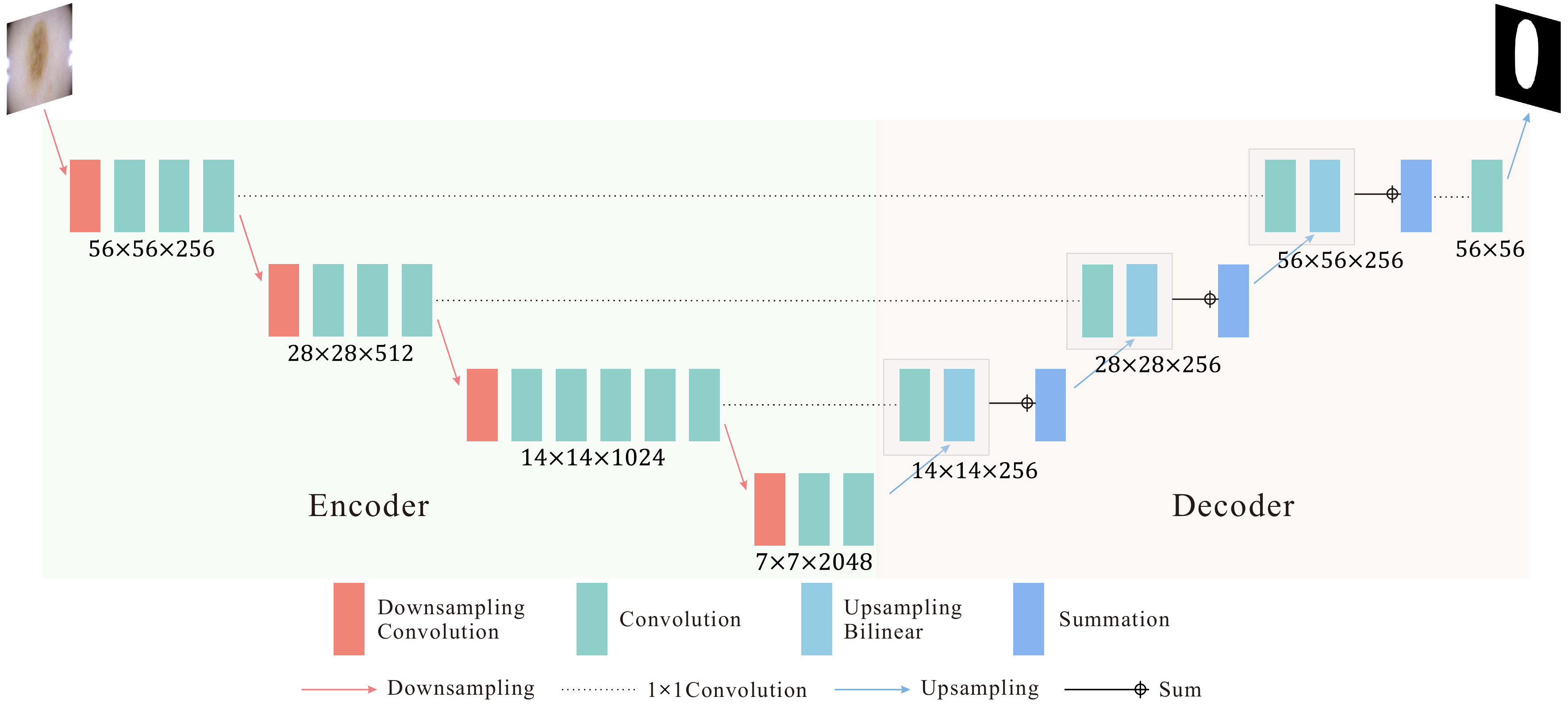}
\caption{Detailed inference structure of skin lesion segmentation network. The encoding and decoding modules are consistent with the learning phase, and the segmentation results are output directly by the binary-thresholding method.}
\label{figure7}
\end{center}
\end{figure*}

In contrast to the fixed structure of ResNet50 used for contrastive learning, the ResNet50+FCN architecture is not standardized across various research. As depicted in Figure \ref{figure7}, we upsample the output of ResNet50's layer $2$, $3$, and $4$ to generate multi-scale features, after which these multi-scale features are merged to produce a $56 \times 56 \times 256$ feature map. This map is subsequently fed into a $1 \times 1$ convolution layer, which retains the spatial dimensions of $56 \times 56$ (i.e., 1/4th of the input image size) while altering the channel depth from 256 to 1. Following a bilinear-interpolation operation, the segmentation output matches the height and width of the input image. During the learning phase, only certain locations contribute to the computation of loss, while uncertain locations are excluded.

\textbf{Iterative-Refinement:} We design an iterative-refinement module. In the module, we define $l_1$, $l_2$, and $l_{n-1}$ to refer to the pseudo-label produced by the 1st, 2nd, and (n-1)th iteration, and $l_0$ to be the initial label defined earlier. Each iteration is set to be 300 epochs and $n$ is empirically set to be 5. Except for the 0th iteration, the other iterations do not need the feature extraction and CCAM, where saliency maps are generated directly by the ResNet50+FCN segmentation network.

\textbf{Uncertainty Loss:}  We introduce an uncertainty loss function, the background is assigned with 0, the foreground is with $1$, and the uncertain locations to be 0.5. For the $i^{th}$ iteration, where $i \in [0, n)$, the definition of the uncertainty loss function is defined in Equation.\eqref{eq3} as the Binary Cross-Entropy (BCE) Loss.

\begin{equation}
Loss = -\frac{(l_i-0.5)^2 \cdot (l_i\log{}{(1-p)}+(1-l_i)\log{}{(1-p)})} {h*w-4*(1-l_i)\cdot l_i^T}
\label{eq3}
\end{equation}
where $l_i$ is the pseudo-label for $i$th iteration, $p$ is the predicted result of the network, and $\cdot$ is the dot product.

\subsection{Model inference} \label{sec:inference}
In the inference phase of Figure \ref{figure2}, both the decoding and encoding networks are used as the same to the learning phase. After loading the optimal parameters of the learning phase, the predicted segmentation results are generated directly using the binary thresholding method without the need for any additional modules. The default threshold is set to 0.5.

\section{Experiments}
\subsection{Datasets}
In order to verify the effectiveness of the proposed method, we conduct experiments on three public data sets, i.e., ISIC-2017 \cite{b20}, ISIC-2018 \cite{b21}, and PH2 \cite{b22} datasets. The ISIC-2017 dataset comprises 2000 training images, and 600 test dermoscopic images. The ISIC-2018 dataset comprises a training set with 2596 annotated dermoscopy images and a test set with 1,000 images. The PH2 dataset consists of 200 dermoscopic images. Each image in these datasets has been professionally annotated as the ground truth. These annotations are employed for the validating the proposed method, rather than for training. We resize all images into a uniform dimension of $224 \times 224$ pixels.

\subsection{Evaluation metrics}
The evaluation methods include Accuracy (ACC in Equation.\eqref{eq4}), Dice Coefficient (Dice in Equation.\eqref{eq5}), Jaccard Index (JAC in Equation.\eqref{eq6}), Sensitivity (SEN in Equation.\eqref{eq7}), Specificity (SPE in Equation.\eqref{eq8}), Recall ( in Equation.\eqref{eq9}), and Precision (Equation.\eqref{eq10}). Recall and SEN have the same formula.  We rank them according to the time of publication. TP, TN, FP, and FN are true positives, true negatives, false positives, and false negatives, respectively. The value is 0 or 1, where 0 represents the background, and 1 represents the foreground. Our method uses a binary-threshold to screen the foreground and background in the comparison phase.

\begin{align}
& ACC=\frac{TP+TN}{TP+TN+FP+FN} \label{eq4} \\
& DIC=\frac{2 \times TP}{2 \times TP+FP+FN} \label{eq5}\\
& JAC=\frac{TP}{TP+FP+FN} \label{eq6}\\
& SEN=\frac{TP}{TP+FN} \label{eq7}\\
& SPE=\frac{TN}{TN+FP} \label{eq8}\\
& Recall=\frac{TP}{TP+FN} \label{eq9}\\
& Precision=\frac{TP}{TP+FP} \label{eq10}
\end{align}

\subsection{Comparison with state-of-the-art methods}
We select 10 state-of-the-art supervised methods and 10 state-of-the-art unsupervised methods for comparison. The relative supervised methods mainly refer to Ms RED\cite{b79} and Wang et al.\cite{b7}, and the relative unsupervised methods mainly refer to SLED\cite{b98}. In this section, we employed five evaluation metrics to gauge the performance of different methods: Accuracy (ACC), Dice Coefficient (Dice), Jaccard Index (JAC), Sensitivity (SEN), and Specificity (SPE).

\begin{table*}[htbp]
\scriptsize
\setlength{\tabcolsep}{9pt}
\centering
\renewcommand\arraystretch{1.4}
\begin{minipage}[c]{0.49\textwidth}
\centering
\captionof{table}{Comparison with $\mathbf{supervised}$ methods in ISIC-2017 datasets}
\label{table1}
\begin{tabular}{l c c c c c }
\hline
   \multirow{2}{*}{Methods} & \multicolumn{5}{c}{Averaged evaluation metrics (\%)}\\\cline{2-6}
& ACC & DIC & JAC  & SEN & SPE\\
\hline
AG-Net \cite{b57} &93.5& 85.3& 76.9& 83.5& 97.4\\
\hline
CE-Net \cite{b59}  & 94.0& 86.5& 78.5& 86.9& 96.4\\
\hline
CPFNet \cite{b66} & 94.2& 86.3& 78.4& 84.4& 97.1\\
\hline
Inf-Net \cite{b67} & 94.2& 86.5& 78.5& 84.8& $\mathbf{97.5}$\\
\hline
DW-HieraSeg \cite{b70} & 94.0& 86.7& 79.0& 87.9& 95.7\\
\hline
PyDiNet \cite{b71} & 94.1& 86.2& 78.1& 84.2& 96.5\\
\hline
SESV-DLab \cite{b72} & 94.1& 86.8& 78.8& 88.3& 95.7\\
\hline
Wang et al.\cite{b7} & $\mathbf{94.6}$ & $\mathbf{87.8}$ & $\mathbf{80.3}$ & $\mathbf{88.6}$ & 96.4\\
\hline
Ms RED \cite{b79} &94.1 & 86.5 &78.6 & - & - \\
\hline
EIU-Net \cite{b96} & 93.7 & 85.5 & 77.1 & 84.2 & 96.8\\
\hline
$\mathbf{Proposed}$ & 90.5 & 80.5 & 68.5 &  $\mathbf{88.6}$ & 93.7\\
\hline
\end{tabular}
\end{minipage}
\begin{minipage}[c]{0.49\textwidth}
\centering
\captionof{table}{Comparison with $\mathbf{unsupervised}$ methods in ISIC-2017 datasets}
\label{table2}
\begin{tabular}{l c c c c c }
\hline
   \multirow{2}{*}{Methods} & \multicolumn{5}{c}{Averaged evaluation metrics (\%)}\\\cline{2-6}
& ACC & DIC & JAC  & SEN & SPE\\
\hline
Sp. Merging \cite{b103} & 79.9& 54.7 & 46.0 & 59.2& 88.7\\
\hline
DRC \cite{b104}  & 83.8 & 59.1 &45.4 & 70.4 &95.6 \\
\hline
Saliency-CCE \cite{b105}  &83.9 & 61.8 &49.5 &74.1 & 93.0 \\
\hline
A2S-v2 \cite{b106} & 82.9& 61.4& 51.1 & 68.7& 92.8\\
\hline
SpecWRSC \cite{b107} & 83.0& 61.0& 50.5& 68.2& 90.0\\
\hline
NCut \cite{b108} & 83.8& 62.1& 51.8 & 67.1& 90.9\\
\hline
K-means \cite{b109} & 84.9& 67.8& 58.1& 70.8& 90.6\\
\hline
SGSCN \cite{b110} & 85.1& 60.0& 50.0& 55.0& $\mathbf{95.6}$\\
\hline
SDI+ \cite{b94}  & 88.8 & 78.2 & $\mathbf{69.2}$ & 81.3 & 92.7 \\
\hline
SLED \cite{b98} & 88.8 & 73.9 & 64.5 & 78.0& 94.6\\
\hline
$\mathbf{Proposed}$ & $\mathbf{90.5}$ & $\mathbf{80.5}$ & 68.5 & $\mathbf{88.6}$ & 93.7 \\
\hline
\end{tabular}
\end{minipage}
\end{table*}

\begin{table*}[htbp]
\scriptsize
\setlength{\tabcolsep}{9pt}
\centering
\renewcommand\arraystretch{1.4}
\begin{minipage}[c]{0.49\textwidth}
\centering
\captionof{table}{Comparison with $\mathbf{supervised}$ methods in ISIC-2018 dataset}
\label{table3}
\begin{tabular}{l c c c c c }
\hline
   \multirow{2}{*}{Methods} & \multicolumn{5}{c}{Averaged evaluation metrics (\%)}\\\cline{2-6}
& ACC & DIC & JAC  & SEN & SPE\\
\hline
FCN \cite{b4} &94.8 & 85.7 &77.5 & 87.2 & 96.2 \\
\hline
U-Net \cite{b5} &94.9 & 86.1 &78.3 & 90.5 & 96.6 \\
\hline
AG-Net \cite{b57} &95.3 & 87.4 &80.0 & 89.2 & 97.3 \\
\hline
CE-Net \cite{b59} &95.8 & 88.2 &81.3 & 91.5 & 96.4 \\
\hline
CPFNet \cite{b66} &96.3 & 90.1 &83.5 & $\mathbf{91.7}$ & 96.6 \\
\hline
Inf-Net \cite{b67} &96.4 & 89.9 &83.2 & 90.8 & 97.1 \\
\hline
PyDiNet \cite{b71} &96.3 & 89.9 &83.2 & 91.4 & 96.6 \\
\hline
Wang et al. \cite{b7} &96.5 & $\mathbf{90.5}$ &$\mathbf{84.3}$ & 91.1 & 97.2 \\
\hline
Ms RED \cite{b79} &96.2 & 90.0 &83.5 & 90.5 & - \\
\hline
EIU-Net \cite{b96} & $\mathbf{96.7}$ & 90.2 & 83.6 & 90.7 & 96.7\\
\hline
$\mathbf{Proposed}$ & 88.5 & 80.8 & 68.3 & 90.9 & 87.8 \\
\hline
\end{tabular}
\end{minipage}
\begin{minipage}[c]{0.49\textwidth}
\centering
\captionof{table}{Comparison with $\mathbf{unsupervised}$ methods in ISIC-2018 dataset}
\label{table4}
\begin{tabular}{l c c c c c }
\hline
   \multirow{2}{*}{Methods} & \multicolumn{5}{c}{Averaged evaluation metrics (\%)}\\\cline{2-6}
& ACC & DIC & JAC  & SEN & SPE\\
\hline
USASOD \cite{b102}  & 67.9 & 24.5 & 18.9 & 79.6 & 24.5 \\
\hline
Sp. Merging \cite{b103} & 84.5& 70.1 & 68.9 & 69.8& 92.8\\
\hline
DRC \cite{b104}  & 83.9 & 68.7 &56.1 & 70.0 &97.0 \\
\hline
Saliency-CCE \cite{b105}  &85.5 &72.9 & 61.9 & 77.4 &94.3  \\
\hline
A2S-v2\cite{b106} & 86.2& 75.0& 65.9 & 72.1& $\mathbf{97.4}$\\
\hline
SpecWRSC \cite{b107} & 81.2& 69.0& 58.4& 69.4& 86.3\\
\hline
NCut \cite{b108} & 82.5& 69.1& 58.8 & 68.1& 88.3\\
\hline
K-means \cite{b109} & 83.7& 71.5& 61.6& 71.7& 88.1\\
\hline
SGSCN \cite{b110} & 82.3& 71.0& 61.8& 71.2& 87.4\\
\hline
SLED \cite{b98} & 86.9 &77.7& $\mathbf{69.4}$ & 80.1  & 91.8\\
\hline
$\mathbf{Proposed}$ & $\mathbf{88.5}$ & $\mathbf{80.8}$ & 68.3 & $\mathbf{90.9}$ & 87.8 \\
\hline
\end{tabular}
\end{minipage}
\end{table*}

\begin{table*}[htbp]
\scriptsize
\setlength{\tabcolsep}{9pt}
\centering
\renewcommand\arraystretch{1.4}
\begin{minipage}[c]{0.49\textwidth}
\centering
\captionof{table}{Comparison with $\mathbf{supervised}$ methods in PH2 datasets}
\label{table5}
\begin{tabular}{l c c c c c }
\hline
   \multirow{2}{*}{Methods} & \multicolumn{5}{c}{Averaged evaluation metrics (\%)}\\\cline{2-6}
& ACC & DIC & JAC  & SEN & SPE\\
\hline
FCN \cite{b4} & 93.5& 89.4& 82.2 & 93.1& 93.0\\
\hline
U-Net \cite{b5}  &- & 87.6 &78.0 &- &- \\
\hline
U-Net++ \cite{b77}  &- & 87.6 &78.0 &- &- \\
\hline
AG-Net \cite{b57} & 93.8& 90.6& 84.0 & 94.7& 94.2\\
\hline
CE-Net \cite{b59} & 94.8& 91.7& 85.2& 95.3& 94.2\\
\hline
FC-DPN \cite{b61} & 93.6& 90.3& 83.5 & 94.7& $\mathbf{96.3}$\\
\hline
CPFNet \cite{b66} & 94.8& 91.8& 85.5& 96.5& 94.3\\
\hline
Inf-Net \cite{b67} & 94.5& 92.0& 85.8& 97.1& 94.1\\
\hline
PyDiNet \cite{b71} & 94.7& 91.9& 85.6& 97.3& 92.5\\
\hline
Wang et al. \cite{b7} & $\mathbf{95.4}$ & $\mathbf{92.5}$ & $\mathbf{86.6}$ & $\mathbf{98.6}$ & 94.7\\
\hline
$\mathbf{Proposed}$ & 92.4 & 88.9 & 80.1 & 93.6 & 93.1 \\
\hline
\end{tabular}
\end{minipage}
\begin{minipage}[c]{0.49\textwidth}
\centering
\captionof{table}{Comparison with $\mathbf{unsupervised}$ methods in PH2 datasets}
\label{table6}
\begin{tabular}{l c c c c c }
\hline
   \multirow{2}{*}{Methods} & \multicolumn{5}{c}{Averaged evaluation metrics (\%)}\\\cline{2-6}
& ACC & DIC & JAC  & SEN & SPE\\
\hline
USASOD \cite{b102}  & 63.2 & 31.6 & 23.7 & 48.2 & 68.0 \\
\hline
Sp. Merging \cite{b103} & 89.1& 82.8& 74.0& 79.6 & 96.0 \\
\hline
DRC \cite{b104}  & 82.6 & 72.4 &59.8 & 69.2 &$\mathbf{97.7}$ \\
\hline
Saliency-CCE \cite{b105}  &84.6 &76.8 & 65.6 & 78.9 &93.8  \\
\hline
A2S-v2\cite{b106} & 85.7& 78.2& 67.2 & 87.6& 91.4\\
\hline
SpecWRSC \cite{b107} & 86.0& 78.1& 68.5& 79.1& 89.7\\
\hline
NCut \cite{b108} & 88.2& 80.2& 70.5 & 76.2& 94.9\\
\hline
K-means \cite{b109} & 91.6& 86.1& 77.8& 83.8& 95.0\\
\hline
SGSCN \cite{b110} & 91.8& 87.9& 80.2& 84.3& 93.8\\
\hline
SLED \cite{b98} & $\mathbf{93.0}$ & $\mathbf{90.3}$ & $\mathbf{83.5}$ & 90.0  & 96.4\\
\hline
$\mathbf{Proposed}$ & 92.4 & 88.9 & 80.1 & $\mathbf{93.6}$ & 93.1 \\
\hline
\end{tabular}
\end{minipage}
\end{table*}

\subsubsection{Experiments on ISIC-2017 dataset}
The segmentation performance on ISIC-2017 is presented in Table.\ref{table1} and Table.\ref{table2}, where a "-" denotes that a particular method was not evaluated on certain criteria.  Our method differs significantly from supervised methods compared to Dice and JAC. However, ACC and SPE are close to supervised methods. For ACC, our approach is only 4.1\% lower than the highest result achieved by the supervised method. The SPE of our method is also competitive with other methods. Notably, our approach achieves a SEN of 88.6\%, outperforming numerous supervised methods.

When compared to unsupervised methods, our approach surpasses the SLED method in ACC 1.7\%, Dice 6.6\%, SEN 4.0\%, and SPE 10.6\%, with only JAC falling behind by a marginal 0.7\%. Our method has a definite advantage over all other unsupervised methods in terms of performance in ISIC-2017.

\subsubsection{Experiments on ISIC-2018 dataset}

Table.\ref{table3} illustrates the segmentation performance on the ISIC-2018 dataset for supervised methods. In terms of the Sensitivity (SEN) metric, our method surpassed several state-of-the-art supervised methods, such as DO-Net, Inf-Net, and Ms RED, and showed minimal discrepancies with other supervised methods. As for the Accuracy (ACC) criterion, the study by \cite{b7} recorded the best result at 96.5\%, whereas our method attained 88.5\%, leaving a difference of 8.0\%. For the Dice Coefficient (DIC) criterion, our method lagged behind the supervised methods by 4.9\%-9.7\%, achieving a performance rate of 80.8\%.

Table.\ref{table4} illustrates the segmentation performance for unsupervised methods. Similar to the results evaluated in ISIC-2017, our method still performs with large advantages in ISIC-2018. The ACC criterion and the Dice criterion outperform the past optimal results by 1.6\%, and 3.1\% respectively. The SEN outperforms the past optimal method by 10.8\%.

\subsubsection{Experiments on PH2 dataset}
Many studies train their models using the ISIC-2017 dataset and test on PH2 dataset. This is due to the small size of PH2, which only consists of 200 samples and is hence challenging to be trained effectively. Moreover, using the crossing-databases allows for the generalization assessment on different methodologies.

The segmentation performance on the PH2 dataset is depicted in Table.\ref{table5} and Table.\ref{table6}. Our method yields better performance, reaching an Accuracy (ACC) of 92.4\%, a Dice Coefficient (DIC) of 88.9\%, a Jaccard Index (JAC) of 80.1\%, a Sensitivity (SEN) of 93.6\%, and a Specificity (SPE) of 93.1\%. It is superior to the supervised methods U-Net and U-Net++. For the unsupervised methods, SLED shows a large improvement in performance relative to the ISIC dataset. Our method outperforms it by 3.6\% in the SEN criterion, and slightly underperforms it in the ACC, DIC, and JAC criteria, but still outperforms the other methods.

\begin{table*}[htbp]
\scriptsize
\caption{Cross-validate the generalization ability of different methods on ISIC-2018 and PH2 datasets.}
\label{table7}
\centering
\renewcommand\arraystretch{1.4}
 \begin{tabular}{ p{4cm} p{0.7cm}  p{0.7cm}  p{0.7cm}  p{0.7cm}  p{0.7cm}  p{0.8cm}  p{0.7cm}  p{0.7cm}  p{0.7cm}  p{0.7cm}  p{0.7cm} }
\hline
   \multirow{2}{*}{Methods} & \multicolumn{5}{c}{ISIC-2018 $\rightarrow$ PH2 (\%)} &  & \multicolumn{5}{c}{PH2 $\rightarrow$ ISIC-2018 (\%)}\\  \cline{2-6} \cline{8-12}
& ACC & DIC & JAC  & Recall &Precision &  & ACC & DIC & JAC  & Recall & Precision\\
\hline
& \multicolumn{11}{c}{Supervised}  \\
\hline
FCN \cite{b4}  &93.9 & 89.4 &81.9 & $\mathbf{97.6}$ & 83.6 & &84.0 & 65.1&55.7 &77.5 &66.4 \\
\hline
U-Net \cite{b5}  &94.1 & 90.7 &83.7 & 97.1 & 85.9 & &76.7 & 67.0&58.0 &85.3 &64.4 \\
\hline
U-Net++ \cite{b77} & 94.3 & 91.2 & 84.4 & 96.7 & 86.9 & & 85.0 & 67.5 & 58.6 & 78.3 & 69.9 \\
\hline
AttU-Net \cite{b88} & 94.2 & 91.1 & 84.6 & 96.7 & 87.1 & & 84.6 & 70.1 & 60.1 & 84.1 & 67.8 \\
\hline
DeepLabv3+ \cite{b89} & 94.6 & 91.3 & 84.9 & 96.3 & 87.7 & & 86.0 & 69.9 & 60.3 & 83.9 & 68.2 \\
\hline
DenseASPP \cite{b90} & 94.5 & 91.6 & 85.2 & 96.5 & 87.9 & & 84.6 & 69.0 & 60.6 & 81.6 & 70.2 \\
\hline
BCDU-Net \cite{b91} & 94.2 & 90.9 & 83.9 & 95.4 & 87.4 & & 78.4 & 67.3 & 58.3 & 86.5 & 64.2 \\
\hline
Focus-Alpha \cite{b92}  & 94.3 & 91.3 & 84.6 & 96.4 & 87.4 &  & $\mathbf{86.3}$ & 69.8 & 60.2 & 82.3 & 69.2 \\
\hline
CE-Net \cite{b59}  & 94.9 & 92.0 & 85.8 & 96.2 & 88.8 &  & 82.5 & 70.0 & 61.3 & 86.1 & 68.0 \\
\hline
CA-Net \cite{b93}  & 93.5 & 90.4 & 83.2 & 94.1 & 87.8 &  & 82.5 & 66.0 & 56.5 & 80.7 & 65.3 \\
\hline
DO-Net \cite{b78}  & 94.5 & 91.6 & 85.4 & 95.4 & 89.1 &  & 82.4 & 69.7 & 60.9 & $\mathbf{88.9}$ & 65.9 \\
\hline
CPF-Net \cite{b66} & 95.2 & 92.2 & 86.2 & 96.8 & 88.7 &  & 85.1 & 71.1 & 61.7 & 85.2 & 69.1 \\
\hline
Ms RED \cite{b79} & $\mathbf{95.3}$ & $\mathbf{92.5}$ & $\mathbf{86.6}$ & 96.7 & $\mathbf{89.2}$ & & 84.4 & $\mathbf{71.1}$ & $\mathbf{62.3}$ & 84.5 & 70.4 \\
\hline
& \multicolumn{11}{c}{Unsupervised}  \\
\hline
$\mathbf{Proposed}$ & 90.1 & 85.7 & 75.2 & 95.2 & 78.5 & & 83.4 & 68.5 & 54.2 & 73.6 & $\mathbf{71.2}$ \\
\hline
 \end{tabular}
\end{table*}

\subsection{Cross-validation on ISIC-2018 and PH2 datasets}
To further validate the generalization ability of the proposed method, we execute cross-validation on the ISIC-2018 and PH2 datasets. The results are depicted in Table.\ref{table7}, where the gap between our method and the supervised methods in the same validation scenario is further diminished. This is particularly evident in the PH2 $\leftarrow$ISIC-2018 validation where our method closely competes with supervised methods. Here, our method achieves an accuracy of 71.2\%, exceeding the best-supervised method by 0.8\%. Furthermore, in terms of the Accuracy (ACC) criterion, our method only deviates by 2.9\% from the optimal supervised method, surpassing the performance of five supervised methods. The performance with respect to the Dice Coefficient (DIC) criterion mirrors that of ACC, outperforming many supervised methods. In the ISIC-2018$ \leftarrow$PH2 validation, while there still exists a gap between our method and the supervised methods, it is considerably smaller compared to direct validation scenarios.

\begin{table}
\setlength{\tabcolsep}{10.5pt}
\caption{Evaluation of different modules in our method.}
\label{table8}
\centering
\renewcommand\arraystretch{1.4}

 \begin{tabular}{l | c c c c c }
 \hline
   \multirow{2}{*}{Methods} & \multicolumn{5}{c}{Averaged evaluation metrics (\%)}\\\cline{2-6}
& ACC & DIC & JAC  & SEN &SPE\\
\hline
$\mathbf{Baseline}$ & 72.1 & 47.4 &32.3 &73.2 &  71.8  \\
\hline
$\mathbf{Model \ 1}$ &  76.6 & 52.9  &40.7 & 82.0 &  78.3 \\
\hline
$\mathbf{Model \ 2}$ & 77.2 & 55.5 &43.4 &82.7 &  82.6 \\
\hline
$\mathbf{Model \ 3}$ & 86.8 & 68.3 &53.2 &62.7 &  $\mathbf{95.4}$ \\
\hline
$\mathbf{Model \ 4}$ & 89.2 & 77.2 &64.0 &86.7 &  90.7 \\
\hline
$\mathbf{Model \ 5}$ & $\mathbf{90.5}$ & $\mathbf{80.5}$ & $\mathbf{68.5}$ & $\mathbf{88.6}$ & 93.7 \\
\hline
 \end{tabular}
\end{table}

\subsection{Ablation study}
In this section, we conduct ablation studies on various components of our proposed USL-Net. The majority of the experimental work is based on the ISIC-2017 dataset. For clearer depiction and understanding, we will denote certain parts of the process using abbreviations. Subsequent subsections will provide a more detailed dissection of the ablation study for each individual process.

$\mathbf{Baseline}$: Using MoCov2 to extract features, we generate CAM by Smooth Grad-CAM++ method, and generate segmentation results by binary-thresholding.

$\mathbf{Model \ 1}$:  Replacing Smooth Grad-CAM++ with CCAM generates the CAM method as a saliency map based on the Baseline.

$\mathbf{Model \ 2}$: Connectivity and centrality detection for UM is added to filter the foreground based on Model 1.

$\mathbf{Model \ 3}$: Learning using pseudo-label in segmentation networks with BCE loss. Pseudo-label is produced by Model 2.

$\mathbf{Model \ 4}$:  Uncertainty self-learning with pseudo-label in segmentation networks with USL loss. Based on the Model 1, the pseudo-label was generated by UM. UM is the proposed uncertainty module.

$\mathbf{Model \ 5}$: Iterative-refinement with USL loss. First, we used the Model 4 results filter by UM as a pseudo-label. Afterward, the result of each learning was used as the next pseudo-labeling by the UM filter.

$\mathbf{Quantitative \ Study:}$ The ablation study incorporates a total of six models. We use MoCov2 with Smooth Grad-CAM++ as the Baseline. The quantitative results are displayed in Table.\ref{table8}. Initially, CCAM is employed in place of Smooth Grad-CAM++. This change improves pseudo-labeling quality by 0.6\%, 2.6\%, 2.7\%, 0.7\% and 4.3\% in ACC, DIC, JAC, SEN, and SPE criteria, respectively. Subsequently, we apply the UM module for connectivity and centrality detection to filter out the foreground. This results in the same improvements in ACC, DIC, JAC, SEN and SPE criteria as with the first modification.
Model 3 directly is learned from Model 2's pseudo-labeling, leading to improvements of 9.6\%, 12.8\%, 9.8\% and 12.8\% for ACC, DIC, JAC and SPE criteria, respectively. However, it receives a decrease of 20.0\% for the SEN criterion.
Our uncertainty self-learning method, implemented in Model 4, improved by 20.0\%, 21.7\%, 20.3\%, 4.0\%, and 8.2\% relative to Model 2 in the respective criteria, showing a substantial overall enhancement. 
Finally, iterative-refinement in Model 5 shows a performance improvement of 1.3\%, 3.3\%, 4.5\%, 1.9\% and 3.0\%, respectively.

\begin{figure}[!h]
\begin{center}
\includegraphics[width=1\linewidth]{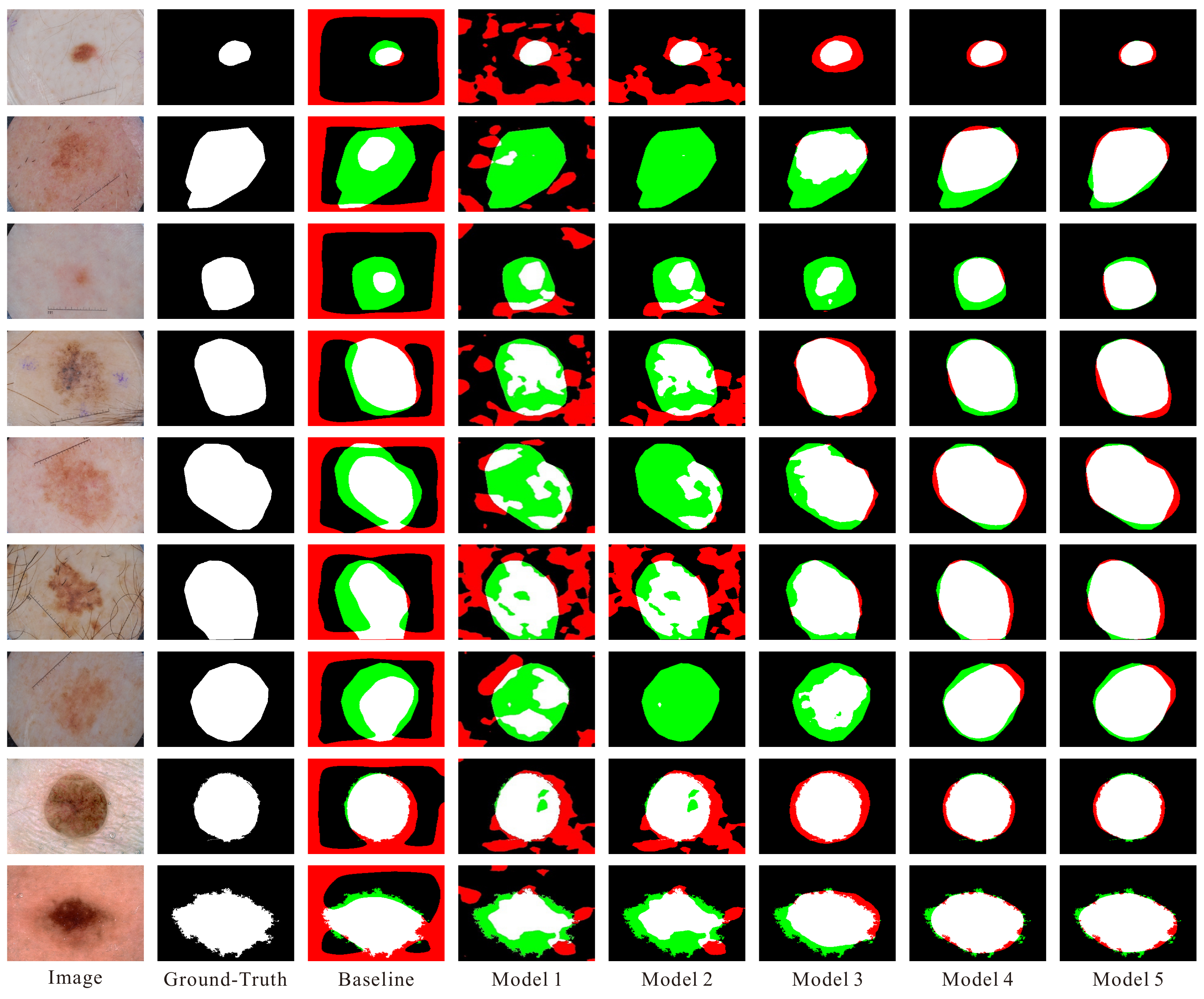}
\caption{Visualization of segmentation results for different models in our framework. Red represents over-segmentation regions and green represents under-segmentation regions, respectively.}
\label{figure8}
\end{center}
\end{figure}

$\mathbf{Qualitative \ Study:} $
Figure \ref{figure8} qualitatively showcases the benefits of the proposed modules for skin lesion segmentation. The Baseline model primarily struggles with over-segmentation of boundary regions and under-segmentation of lesion regions, finding it challenging to distinguish between these regions. Model 1 generates more noise in the background region; however, this can be mitigated by the application of connectivity detection and centrality detection in Model 2. Model 3, which uses direct supervised learning with the results of Model 2 as pseudo-labels, significantly reduces background and edge noise and more accurately identifies lesion locations. However, it still produces instances of over-segmentation and under-segmentation. Our proposed method in Model 4 and the iterative-refining technique in Model 5 display improved results, successfully alleviating both over-segmentation and under-segmentation.

\begin{table}[h]
\setlength{\tabcolsep}{6pt}
\caption{Evaluation of different contrastive learning structures in our method.}
\label{table9}
\centering
\renewcommand\arraystretch{1.4}

 \begin{tabular}{l | p{0.1cm} p{0.1cm}  c | c c c c c }
 \hline
   \multirow{2}{*}{Methods} & \multirow{2}{*}{$S_t$} &  \multirow{2}{*}{$S_c$} &   \multirow{2}{*}{$C_c$} & \multicolumn{5}{c}{Averaged evaluation metrics (\%)}\\\cline{5-9}
&  & & & ACC & DIC & JAC & SEN & SPE\\
\hline
$MoCov3$ &\checkmark & & & 74.7 & 25.1 &19.1 &22.3 &  90.6 \\
\hline
$DINO$ &\checkmark &  & & $\mathbf{76.9}$ & 22.0 &16.1 &18.8&95.1   \\
\hline
$DINOv2$ &\checkmark  & & & 76.3 & 0.3 &0.2 &0.2 &  $\mathbf{99.8}$ \\
\hline
$SimCLR$ & & \checkmark &  & 70.3 & 46.3 &31.6 &80.2 &  73.2 \\
\hline
$MoCov1$ & &\checkmark  & & 68.3 & 45.1 &30.5 &73.0&  69.4 \\
\hline
$MoCov2$ & &\checkmark & & 72.1 & 47.4 &32.3 &73.2 &  71.8 \\
\hline
$SimCLR$ & & & \checkmark & 74.7 & 51.6 &39.7 &80.4 &  75.8 \\
\hline
$MoCov1$ & & & \checkmark & 73.1 & 49.5 &36.8 &78.9&  74.2 \\
\hline
$MoCov2$ &  & & \checkmark& 76.8 & 52.4 &40.5 &81.0 &  78.7 \\
\hline
$Fused_{C_c}$ &  & & \checkmark &  $\mathbf{76.9}$ & $\mathbf{52.9}$  &$\mathbf{40.7}$ & $\mathbf{82.0}$ & 78.3 \\
\hline
 \end{tabular}
\end{table}

\subsubsection{Comparisons of different contrastive learning and CAM methods}

In this sub-section, we provide a detailed evaluation of the performance of multiple contrastive learning methods, as well as the CCAM. Our method extracts features with the fusion of multiple contrastive learning methods, so we select multiple contrastive learning methods for experiments and the optimal three methods for fusion. Additionally, we compare the CCAM with the classical Smooth Grad-CAM++ method. The specific experimental results are shown in Table \ref{table9}, where $S_t$ denotes Smooth Grad-CAM++ with Transformer structure, $S_t$ denotes Smooth Grad-CAM++ with ResNet50 structure, and $C_c$ represents CCAM with ResNet50 structure.

The Transformer-based method contains MoCov3 \cite{b111}, DINO \cite{b112} and DINOv2 \cite{b116}. The ResNet50-based method includes SimCLR \cite{b39}, MoCov1 \cite{b40}, and MoCov2 \cite{b41}. From the experimental results, the Transformer structure combined with Smooth Grad-CAM++ does not perform well compared to the ResNet50 structure. When the same contrastive learning method is used for feature extraction, a significant performance difference exists between Smooth Grad-CAM++ and CCAM. Due to the inferior quality of Smooth Grad-CAM++, the fusion of the generated CAM does not enhance the performance. In contrast, the fusion of CAMs generated by CCAM improves the quality of the pseudo-labels.

\begin{figure}[htbp]
\begin{center}
\includegraphics[width=0.9\linewidth]{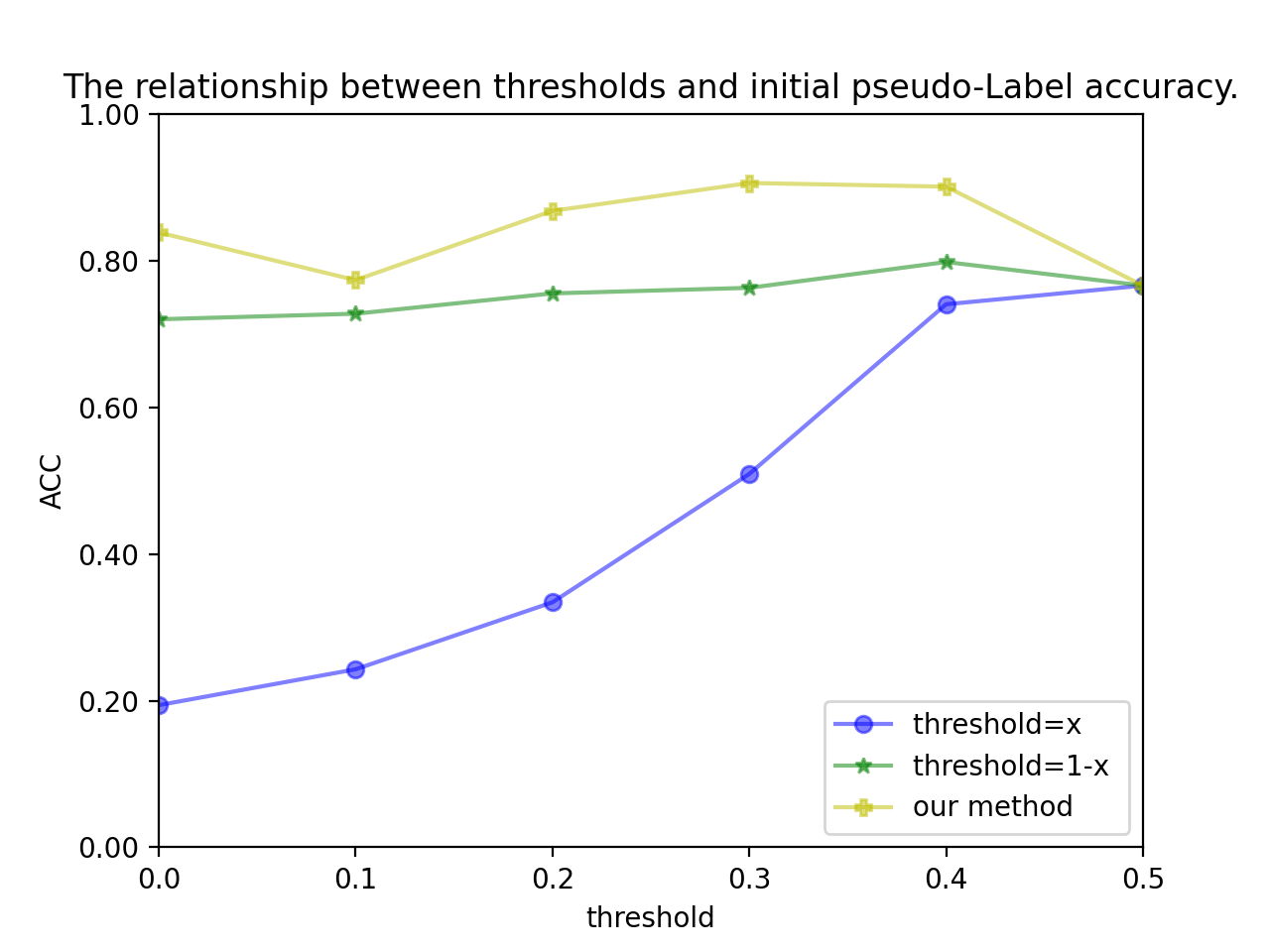}
\caption{Comparison of the accuracy of initial pseudo-labels with the binary-threshold methods and our method.}
\label{figure10}
\end{center}
\end{figure}

\subsubsection{Evaluating the performance of UM and different iterations}

Figure \ref{figure10} illustrates the superiority of the Uncertainty Module (UM) over the binary-threshold method. As depicted in the figure, the blue line corresponds to a threshold of $x$ and the green line represents a threshold of $1 - x$, $x$ ranges from $0$ to $0.5$. In the proposed method, values below $x$ are considered as the background, values above $1-x$ are considered as the foreground, and the range between $(x, 1-x)$ is defined as the uncertainty region. The overall qualities of pseudo-labeling benefit from UM and show stable performance in a large range.

\begin{figure}[h]
\begin{center}
\includegraphics[width=1\linewidth]{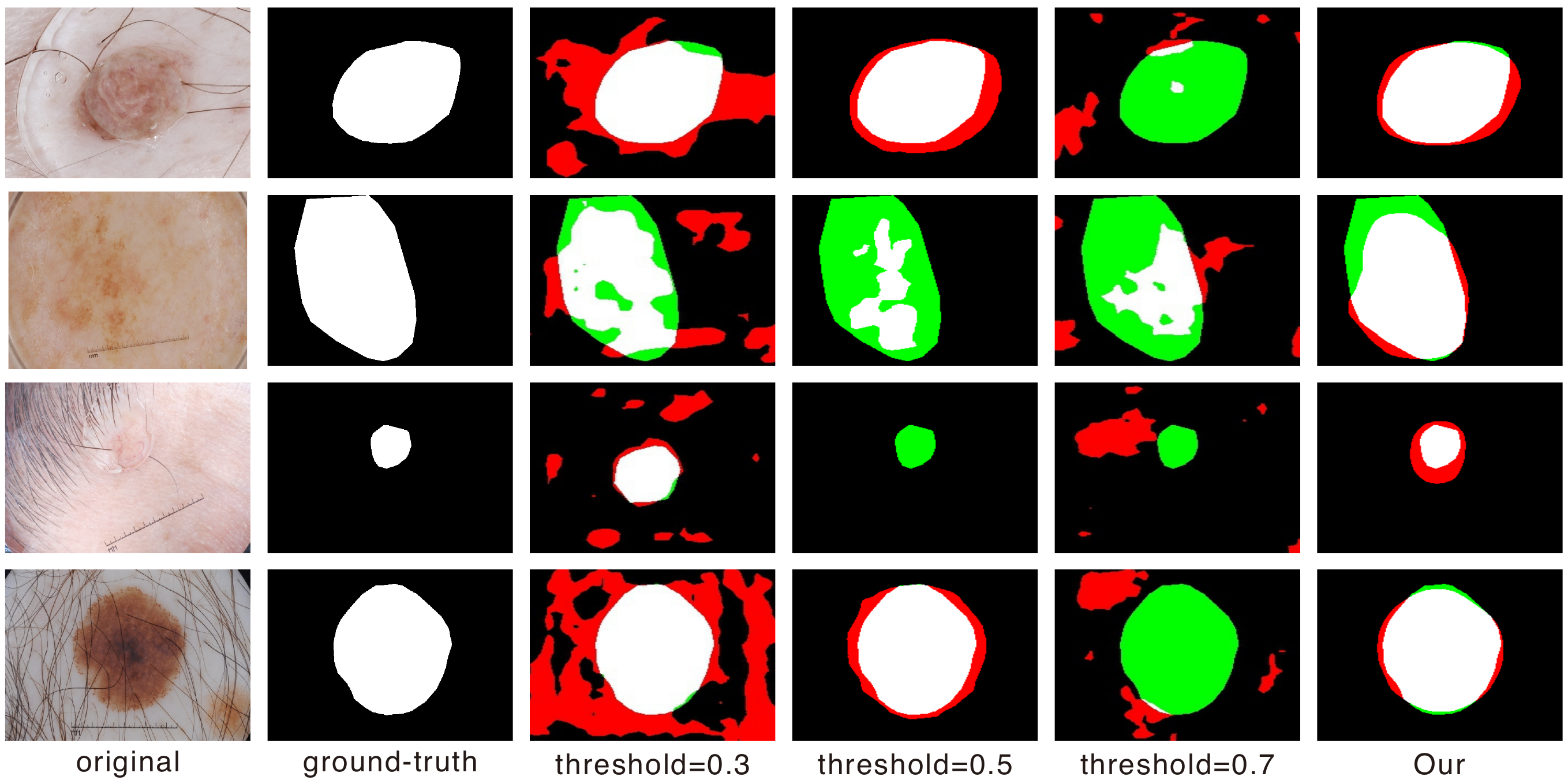}
\caption{A qualitative comparison of the quality of pseudo-labeling produced by the threshold method and the uncertainty region setting method, both consistently utilizing only one iterative learning process.}
\label{figure9}
\end{center}
\end{figure}

Figure \ref{figure9} qualitatively demonstrates the benefits of the Uncertainty Module (UM). We compare the results generated by thresholds of 0.3, 0.5 and 0.7 using a single iterative learning step. When the threshold is set to be 0.3, over-segmentation occurs more frequently. Conversely, when the threshold is set to be 0.7, a high number of results are under-segmented. A threshold of 0.5 somewhat balances over-segmentation and under-segmentation, but our method still outperforms it. In a dataset, varying thresholds may lead to errors in different data samples, and such errors are difficult to avoid.

Furthermore, Table.\ref{table10} displays the results of different iterations in the refinement process. $Model \ 4$ represents the results based on the initial pseudo-label learning, and $Model \ 5_1$ to $Model \ 5_4$ represent the results of $Model \ 5$ from the $1^{st}$ to the $5^{th}$ iterations, respectively. The optimal results are achieved in the second iteration $Model \ 5_2$, which is used to compare with other methods in the experiments.

\begin{table}[hp]
\setlength{\tabcolsep}{10pt}
\caption{Evaluation results in different iterations for the proposed method.}
\label{table10}
\centering
\renewcommand\arraystretch{1.4}

 \begin{tabular}{l | c c c c c }
 \hline
   \multirow{2}{*}{Methods} & \multicolumn{5}{c}{Averaged evaluation metrics (\%)}\\\cline{2-6}
& ACC & DIC & JAC  & SEN &SPE\\
\hline
$Model \ 4$ & 89.2 & 77.2 &64.0 &86.7 &  90.7 \\
\hline
$Model \ 5_1$ & 90.3 & 78.2 & 65.5 & 83.7 &  92.9 \\
\hline
$Model \ 5_2$ & 90.5 & $\mathbf{80.5}$ & $\mathbf{68.5}$ & $\mathbf{88.6}$ & $\mathbf{93.7}$ \\
\hline
$Model \ 5_3$ & $\mathbf{90.9}$ & 80.4 &68.3 & 86.3 & 93.3 \\
\hline
$Model \ 5_4$ & 90.8 & 80.1 &67.9& 86.2  & 93.1 \\
\hline
 \end{tabular}
\end{table}

\section{Discussion}
Due to the inherent noise in dermatoscopy images, traditional methods can not easily achieve satisfactory results. In unsupervised segmentation methods, pseudo-labels are generated to guide the training of networks. We aim to enhance the performance of unsupervised skin lesion segmentation by reducing the uncertainty caused in the saliency maps and improving the quality of pseudo-labels.

We perform rigorous validation across multiple datasets, including ISIC-2017, ISIC-2018 and PH2, which highlights its universal applicability and noise resistance for skin lesion segmentation. Efforts have been especially put on analyzing the saliency features of the lesion areas, which can provide more robust evidence for pseudo-label learning. Through the iterative optimization process, the segmentation results are refined, therefore it eliminates the need to develop separate segmentation models for different lesion types. Free from re-training with new labeled samples for adapting to various scenes, USL-Net can be integrated into a software platform and deployed on the server side, facilitating the clinic inspection.

%

Although the notable improvements in the unsupervised segmentation, the proposed method still can not outperform most supervised methods. Due to the complex context information in challenging scenarios such as purple iodophor regions and hair occlusions, USL-Net does not completely avoid the failures in learning the pseudo-labels for salient regions. Furthermore, the proposed network relies on selecting an optimal saliency threshold (0.35-0.65) for filtering uncertainty regions, which makes it less generalized. The structure of USL-Net is relatively complex, incorporating multiple modules i.e., contrastive learning, CCAM, UM, and iterative-refinement. Although each module is designed independently, allowing the network to be customized and upgraded as necessary, it may be further compressed for reuse in practical implementation and deployment.

In the future, we plan to improve the proposed method by incorporating the semantic knowledge such as objects shape or spatial context from the lesion images. In order to bring the network up to the requirement for practical applications, unsupervised saliency detection and other related techniques can be adopted to handle various scenarios effectively. Through these enhancements, we can bridge the performance gap and further improve the applicability of the USL-Net framework.

\section{Conclusions}
In conclusion, this paper addressed the challenges posed by the intensive labeling in the supervised methods and the limited performance of the unsupervised methods in skin lesion segmentation scenarios. We proposed an innovative Uncertainty Self-Learning Network (USL-Net), which incorporates three contrastive learning methods for feature extraction and employs the CCAM method to generate saliency maps. The saliency maps are then used to produce pseudo-labels instead of relying on ground-truth labels for iterative learning. To enhance the quality of pseudo-labels and mitigate the noise, we introduced an Uncertainty Module (UM) with the self-learning technique to reduce pseudo-labeling errors. Extensive experiments were conducted on three publicly available datasets. The results showed that the proposed USL-Net significantly narrows the performance gap between the unsupervised and the supervised methods while eliminating the need for manual labeling. Our approach achieved state-of-the-art performance compared with other unsupervised methods. Moreover, the ablation study performed on the ISIC-2017 dataset further confirmed the substantial improvement obtained from uncertainty self-learning compared to other learning approaches.

\section*{Declaration of Competing Interest}
The authors declare that we have no known competing financial interests or personal relationships that could have appeared to influence the work reported in this paper.

\section*{CRediT authorship contribution statement}
$\mathbf{Xiaofan \ Li}$:  Conceptualization, Methodology, Writing-original draft, Writing-reviewer \& editing. $\mathbf{Bo \ Peng}$: Writing-original draft, Writing-reviewer,Funding acquisition, Project administration. $\mathbf{Jie \ Hu}$: Writing-reviewer, Funding acquisition, Project administration. $\mathbf{Changyou \ Ma}$: Funding acquisition, Project administration. $\mathbf{Daipeng \ Yang}$: Writing-original draft.  $\mathbf{Zhuyang \ Xie}$: Conceptualization, Methodology.

\section*{Acknowledgments}
This work was supported by the Natural Science Foundation of Sichuan (No. 2022NSFSC0502), Sichuan Science and Technology Program (No. 2023YFG0354), the Key Research and Development Program of Sichuan Province (No. 2023YFG0125), and Open Research Fund Program of Data Recovery Key Laboratory of Sichuan Province (No. DRN2304).

\bibliographystyle{unsrt}
\bibliography{cas-refs.bib}

\end{document}